%% file: main.tex
\documentclass[10pt,twocolumn,letterpaper]{article}

\usepackage{iccv}              


\definecolor{iccvblue}{rgb}{0.21,0.49,0.74}
\usepackage[pagebackref,breaklinks,colorlinks,allcolors=iccvblue]{hyperref}

\def\paperID{15} 
\def\confName{ICCV}
\def\confYear{2025}

\title{Video Text Preservation with Synthetic Text-Rich Videos }

\author{%
Ziyang Liu \quad Kevin Valencia \quad Justin Cui\\
University of California, Los Angeles \\
{\tt\small \{jackieliu2025, kevinval04, justincui\}@ucla.edu}
}

\begin{document}
\maketitle

\input{sec/0_abstract}    
\input{sec/1_intro}
\input{sec/2_related_works}

\input{sec/3_finalcopy}

{
    \small
    \bibliographystyle{ieeenat_fullname}
    \bibliography{main}
}


\end{document}

%% file: sec/0_abstract.tex
\begin{abstract}

While Text-To-Video (T2V) models have advanced rapidly, they continue to struggle with generating legible and coherent text within videos. In particular, existing models often fail to render correctly even short phrases or words and previous attempts to address this problem are computationally expensive and not suitable for video generation. In this work, we investigate a lightweight approach to improve T2V diffusion models using synthetic supervision. We first generate text-rich images using a text-to-image (T2I) diffusion model, then animate them into short videos using a text-agnostic image-to-video (I2v) model. These synthetic video-prompt pairs are used to fine-tune Wan2.1, a pre-trained T2V model, without any architectural changes. Our results show improvement in short-text legibility and temporal consistency with emerging structural priors for longer text. These findings suggest that curated synthetic data and weak supervision offer a practical path toward improving textual fidelity in T2V generation.
\end{abstract}

%% file: sec/1_intro.tex
\section{Introduction}
\label{sec:intro}

Diffusion models, since its inception, have become the primary method for text-to-image and text-to-video generation. Despite these advancements, however, text-to-video diffusion models still struggle in generating legible text. Generated images and videos often contain unnecessary and impurities in the letters of the rendered text, leading to a decrease in overall quality. Attempts have been made to remedy this issue for text-to-image diffusion models, using various techniques such as using a specific text encoder to encode rendered text for generation \cite{liu2024glyph, liu2024glyph2}, large language models to encode positional and textual information \cite{chen2024textdiffuser}, and fine-tuning specifically on images with rendered text \cite{zhao2024harmonizing}, among many other techniques. These techniques have varying levels of success in improving quality of rendered text in images.

However, there have not been much exploration in techniques to improve text rendering for text-to-video diffusion models. Due to how computationally expensive text-to-video diffusion models are to train and run inference, many of the techniques used to improve the quality of rendered text in text-to-image diffusion models are infeasible for text-to-video models. Our aim is to explore less computationally expensive approaches to improve text-to-video diffusion models in text rendering. Our approach to this problem is by fine-tuning a text-to-video diffusion model over a curated dataset of videos generated by OpenAI's Sora which contain coherent text without artifacts and their corresponding prompts. 

\textbf{Our contribution:} From our experiments, we find that it suffices to fine-tune existing text-to-video models on a curated dataset of videos featuring text and prompts describing the video and the text within. By fine-tuning our text-to-video diffusion model, we are able to produce videos which are able to generate text which are coherent and free from artifacts.

%% file: sec/2_related_works.tex
\section{Related works}
\label{sec:Related works}

\subsection{Diffusion Models for Image and Video Generation}
Diffusion models have been a powerful generative modeling framework, dominating fields like image and video generation. Starting from Denoising Diffusion Probabilistic Models (DDPMs)\cite{ho2020denoising, ho2022video} , they have been successfully applied to both image and video generation. In text-to-image (T2I) generation, models like GLIDE, DALL·E 2, and Stable Diffusion\cite{Singer2022makeavideo} produce high-resolution images aligned with text prompts and can often render legible text when guided carefully. Their ability to preserve high-frequency details makes them a natural starting point for tasks requiring precise spatial structure.

Recent extensions of diffusion to text-to-video (T2V) generation include zero-shot pipelines such as Text2Video-Zero and ControlVideo, and end-to-end models like Video Diffusion Models (VDM), Imagen Video, and Make-A-Video\cite{khachatryan2023text2video, zhang2023controlvideo, ho2022video, Singer2022makeavideo}. While these methods achieve realistic motion and temporal consistency, they frequently fail to generate accurate or readable text across frames. In parallel, identity preservation has emerged as a key benchmark for visual consistency in video generation. Methods like CustomVideo and ConsisID\cite{yu2023customvideo, yuan2025consisid} aim to preserve subject-specific features, often using frequency-aware designs.

We build on Wan2.1\cite{wan21}, which achieves state-of-the-art identity preservation through enhanced temporal modeling and attention-based consistency. Although not designed specifically for text, Wan2.1’s ability to retain high-frequency details across frames makes it a strong foundation for improving textual fidelity in generated videos.

\subsection{Diffusion for text rendering}

Recent work has addressed the challenge of preserving readable text in image generation by introducing structure-aware and OCR-guided enhancements. ARTIST\cite{zhang2024artist} proposes a two-stage framework where a specialized "text diffusion model" first generates text layouts at the glyph level, which are then used to guide a visual diffusion model to produce high-quality, legible text within realistic scenes. This separation of layout and appearance helps prevent artifacts and improves readability. TextDiffuser\cite{NEURIPS2023} also adopts a staged approach: a transformer-based layout planner predicts character positions, and a diffusion model renders images conditioned on these layouts. It introduces a glyph-level OCR loss and trains on MARIO-10M, a large-scale dataset curated specifically for visual text generation. In contrast, AnyText\cite{tuo2024anytext} focuses on multilingual and editable text rendering by incorporating text control tokens and perceptual losses derived from OCR feedback. It leverages an auxiliary latent space to enhance stroke-level accuracy, resulting in more precise and flexible text generation across diverse scripts and fonts. Together, these methods demonstrate that combining structural planning with OCR-aware supervision is effective in improving the clarity and correctness of generated text in images.

While significant progress has been made in preserving readable text in image generation through OCR-guided training, layout planning, and glyph-level losses, these strategies have not yet been widely adopted or explored in the context of video generation. Current text-to-video (T2V) models primarily focus on visual realism and temporal coherence, often at the cost of structured text fidelity. To our knowledge, there is no existing work that explicitly targets the problem of rendering accurate and consistent text across frames in T2V. Our work aims to bridge this gap by adapting a high-performing identity-preserving video model to address the unique challenges of text consistency and legibility in generated videos.

\subsection{Fine-tuning diffusion models}

Due to the computational cost of training text-to-video diffusion models, a common approach to tailor text-to-video diffusion models for certain tasks is to take a pre-trained text-to-video diffusion model's weights, and given a small, curated dataset of videos and prompts, train the model to reproduce the video given the prompt. Previous works have been done in fine-tuning text-to-image diffusion models for tasks such as better resemblance to sample images (ID Preservation) \cite{ruiz2023dreambooth}, fairness \cite{shen2023finetuning}, improving image quality \cite{yang2024using, yuan2024self}, among other tasks. Fine tuning text-to-image and text-to-video diffusion models allow for a much less computationally expensive alternative to 

There are various approaches to fine-tuning diffusion models, varying from strictly reproducing the output provided given a prompt, or using rewards to fine-tune the diffusion models, using directly defined rewards \cite{clark2023directly} or reward models \cite{wallace2024diffusion} or other metrics such as human feedback or self-play \cite{yang2024using, yuan2024self}. 

Further improvements to computational efficiency can be made by using Low-Rank Adaption (LoRA) \cite{hu2022lora, smith2024lora, huang2024context}. Initially designed for large language models, LoRA allows for more parameter-efficient fine tuning by approximating the changes to weight matrices of diffusion models with the product of two matrices of much smaller rank, which significantly reduces the number of trainable parameters and enables much faster fine-tuning.

Another method to improve is FlowGRPO\cite{liu2025flow}, which introduces online reinforcement learning to flow-matching diffusion models by converting deterministic ODEs into stochastic SDEs, enabling exploration-driven fine-tuning via Group Relative Policy Optimization (GRPO). This approach significantly improves compositional accuracy and text fidelity in image generation. Extending FlowGRPO to text-to-video models could enhance temporal consistency and better preserve textual details across frames—tackling a key challenge in current T2V systems.

\begin{figure}[t]
  \centering
  \begin{minipage}[b]{0.4\linewidth}
    \includegraphics[width=\linewidth]{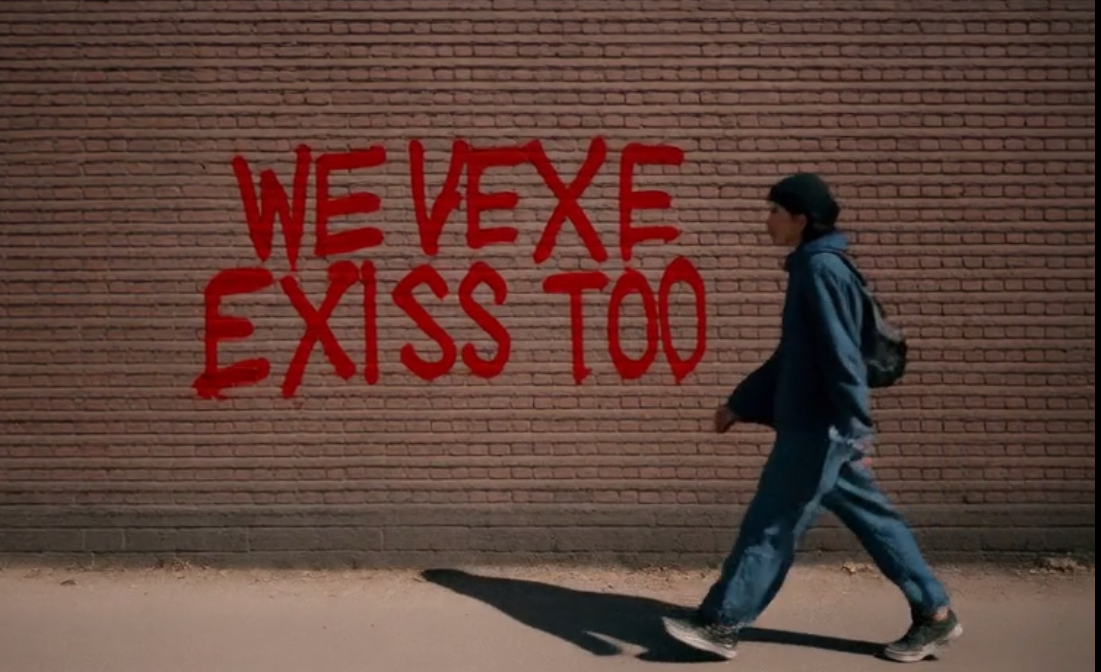}
    
  \end{minipage}
  \hspace{0.05\linewidth}
  \begin{minipage}[b]{0.5\linewidth}
    \includegraphics[width=\linewidth]{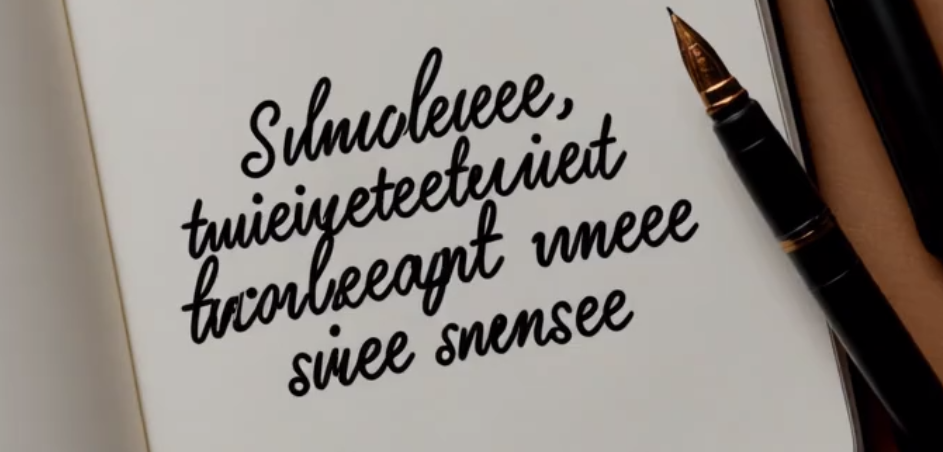}
    
  \end{minipage}
  \caption{Model still struggles with long/complicated tasks. However, it is able to capture the structure of the texts}
  \label{fig:comparison}
\end{figure}

%% file: sec/3_finalcopy.tex
\section{Methods}

\begin{figure*}[t]
  \centering
  \begin{minipage}[b]{0.125\linewidth}
    \includegraphics[width=\linewidth]{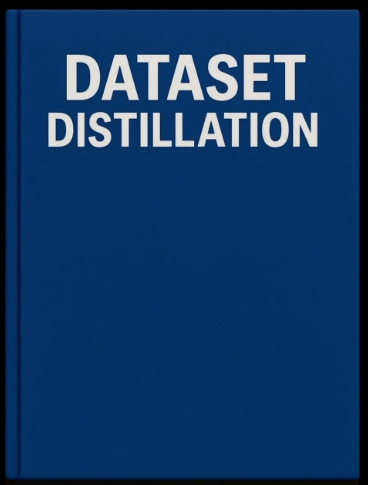}
    \caption*{A book about Distillation}
  \end{minipage}
  \hfill
  \begin{minipage}[b]{0.15\linewidth}
    \includegraphics[width=\linewidth]{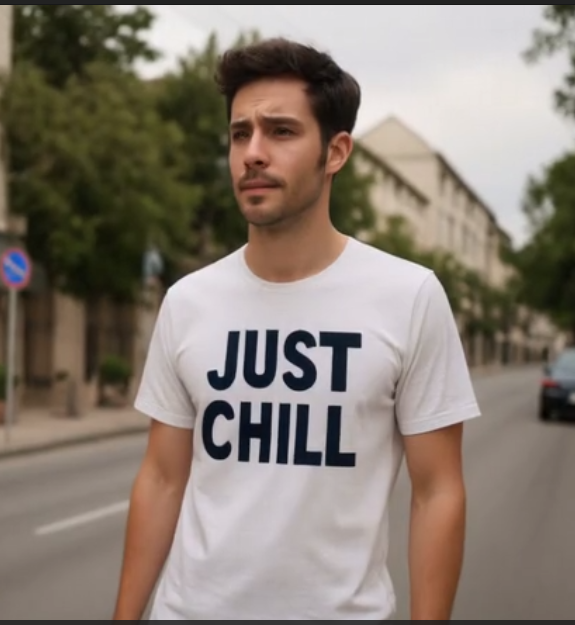}
    \caption*{A chill guy walking on the street}
  \end{minipage}
  \hfill
  \begin{minipage}[b]{0.31\linewidth}
    \includegraphics[width=\linewidth]{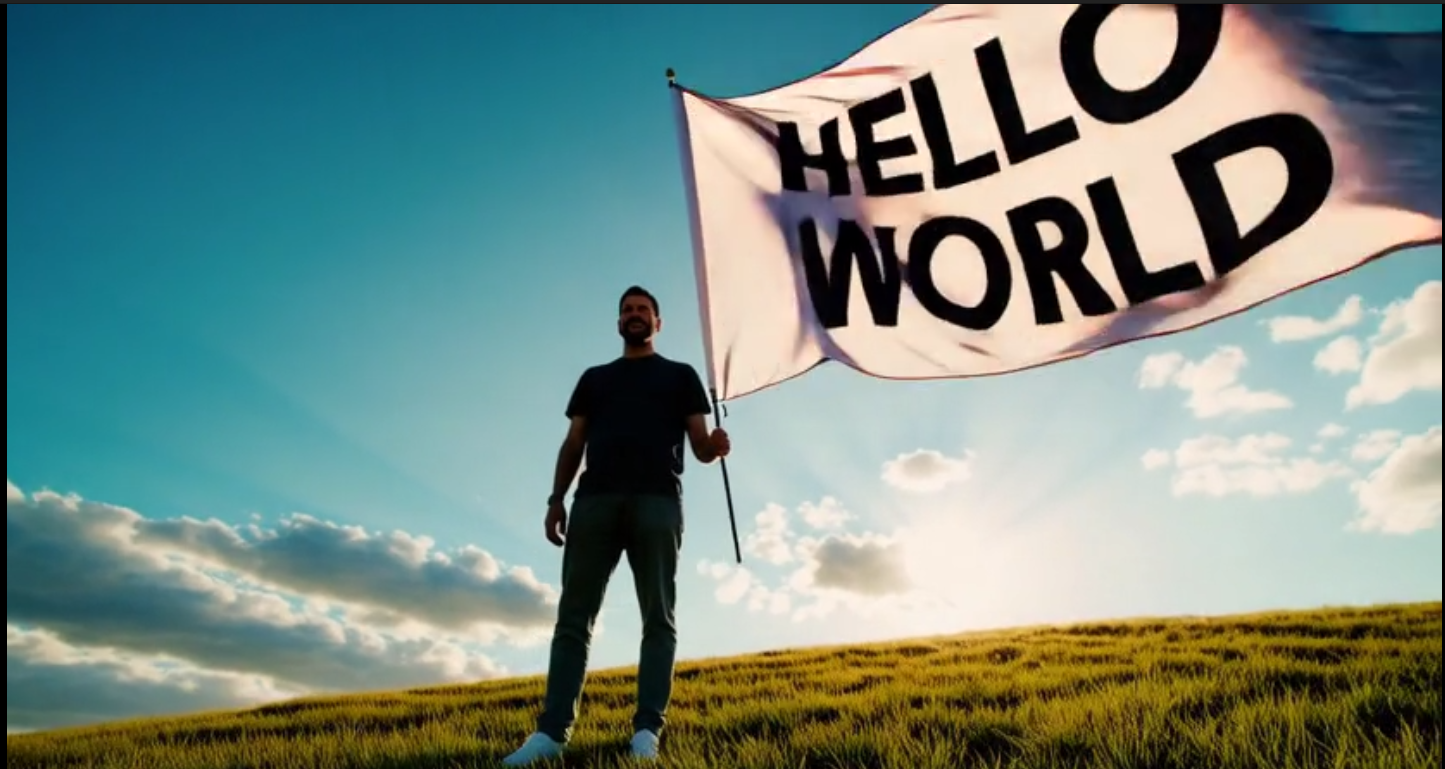}
    \caption*{A man holding Hello World flag on the grass.}
  \end{minipage}
  \hfill
  \begin{minipage}[b]{0.28\linewidth}
    \includegraphics[width=\linewidth]{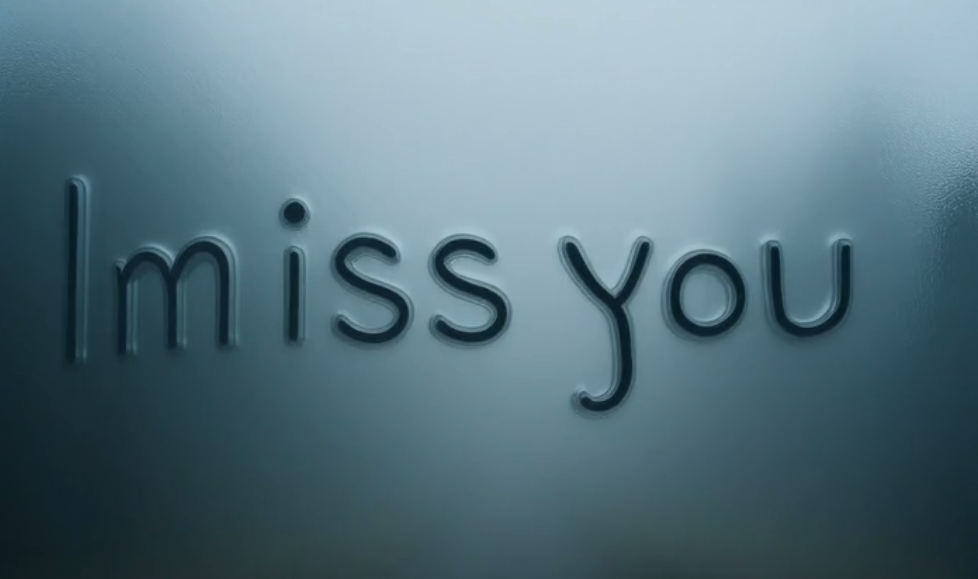}
    \caption*{I miss you on froggy glass gradually showed up}
  \end{minipage}
  \caption{Some example for easy tasks}
  \label{fig:qual_results}
\end{figure*}
\subsection{Data Gathering}
To learn how and when text artifacts emerge in text-to-video (T2V) pipelines, we construct a controlled dataset using a two-stage process. First, we use text-to-image (T2I) diffusion models (e.g., Stable Diffusion) to synthesize high-quality images containing legible and stylistically diverse textual content (e.g., flags, banners, product labels). Unlike T2V models, T2I models are generally able to preserve textual integrity and layout when guided with appropriate prompts (e.g., “A man is standing on a grassy field, holding a flag with 'HELLO WORLD' written on it. The flag is swinging gently in the wind.”).

Next, we treat the generated images as inputs to a "text-free" image-to-video (I2V) synthesis stage. Specifically, we employ pretrained I2V models to generate short video clips where the first frame is the T2I image. Importantly, we do not include any reference to “text” or similar keywords in the prompts provided to the I2V model. In our preliminary experiments, we found that explicitly mentioning text (e.g., "a man wearing a 'JUST CHILL' hoodies is walking on an empty street") often caused I2V models to hallucinate unrealistic or degraded text, overriding the input image. Omitting text-related terms allows the I2V model to focus on temporal smoothness without injecting new textual artifacts.

This setup enables us to isolate how existing I2V models handle spatial high-frequency features (e.g., strokes and letterforms), and provides a benchmarkable setting for measuring text preservation.

\subsection{Fine-Tuning Wan2.1 for Text Preservation}
We adopt Wan2.1-T2V-1.3B \cite{wan21} as our base model for fine-tuning due to its state-of-the-art performance on identity preservation benchmarks in text-to-video generation. Although Wan2.1 was originally developed to maintain the visual consistency of human subjects across frames, its architectural strengths—such as high-frequency-aware feature processing and temporally consistent generation—are well-aligned with the requirements of preserving textual fidelity. In particular, readable text in video frames requires accurate spatial rendering and coherence over time, similar to the needs of facial or object identity preservation. We hypothesize that these strengths make Wan2.1 a strong starting point for addressing the underexplored issue of text degradation in T2V models. Building on this foundation, we fine-tune Wan2.1 using our constructed dataset (Section 3.1) without modifying its loss functions or architecture.
\section{Evaluation and Discussion}
Evaluating text fidelity in generated videos is still an open challenge. Unlike image generation, where OCR tools can be reliably applied, text in video often suffers from frame-wise inconsistency, distortion, and low resolution, making standard metrics unreliable. To better understand how well our model preserves text, we focus on qualitative analysis and structured human observations, rather than relying solely on automated metrics.

We observe that our fine-tuned Wan2.1 model excels at rendering short and simple text phrases (e.g., “Just Chilling”), maintaining legibility across frames. In contrast, most baseline models, including those with state-of-the-art generation quality such as Sora, often fail to produce valid characters—frequently hallucinating symbols or smudged glyphs that bear little resemblance to real text.

Interestingly, for longer sentences, our model does not consistently preserve full word integrity. However, it still demonstrates a remarkable ability to lay out plausible letterforms in sequence, giving rise to coherent “text-like” patterns across the frame. This behavior is analogous to early-stage language models like GPT-2, which could mimic sentence shapes before fully mastering syntax or semantics. We interpret this as a sign that the model has learned structural priors for typography, even when exact decoding fails.

Given the absence of standardized metrics for this task, we advocate for future work to develop text-focused video benchmarks, possibly combining OCR reliability, temporal consistency, and human readability scoring. In the meantime, our evaluation offers both visual comparisons and descriptive insights as a first step toward measuring textual quality in generative video models.

\section{Conclusion}
In this work, we address the persistent challenge of generating legible text in T2V diffusion models. We present a lightweight pipeline that constructs a synthetic dataset by combining T2I and I2V models to generate videos that contain clean and coherent text. Fine-tuning Wan2.1, a pre-trained T2V model on this dataset, leads to significant improvements in short-text legibility and temporal consistency without requiring architectural changes or costly retraining. We also observe the emergence of structural priors for longer text, suggesting that the model is learning layout and typographic patterns even when exact decoding fails. These results demonstrate the potential of weakly supervised synthetic data pipelines to improve textual fidelity in video generation. Future work may explore more diverse text scenarios, real-world data, and the development of standardized benchmarks to evaluate text quality in these models.